%% file: naaclhlt2019.tex
\documentclass[11pt,a4paper]{article}
\RequirePackage{authblk}
\usepackage[hyperref]{naaclhlt2019}
\usepackage{times}

\usepackage{url}
\usepackage{latexsym}

\usepackage{algorithm2e}
\usepackage{amsfonts}
\usepackage{amsmath}
\usepackage{amssymb}
\usepackage{array}
\usepackage{booktabs}
\RequirePackage{pgfkeys,tikz}
\usepackage{ctable}
\usepackage{epstopdf}
\usepackage{etex}
\PassOptionsToPackage{pdftex}{graphicx}
\usepackage{listings} 
\usepackage{comment}
\usepackage{microtype}
\usepackage{multicol}
\usepackage{multirow}
\usepackage{paralist} 
\usepackage{pifont}
\usepackage{ragged2e}
\usepackage{setspace}
\usepackage{titletoc}
\usepackage[normalem]{ulem}
\PassOptionsToPackage{hyphens}{url}
\usepackage{wrapfig}
\usepackage{xcolor}
\usepackage{xspace}
\usepackage{subcaption}
\usepackage{float}

\RequirePackage{flushend}
\usetikzlibrary{shapes,positioning}
\RequirePackage[skip=.5ex plus 1ex]{caption}
\usepackage[backgroundcolor=yellow!30,linecolor=red, bordercolor=red,textsize=scriptsize]{todonotes}
\RequirePackage{adjustbox,array,multirow,colortbl}
\colorlet{newchanges}{red}

\aclfinalcopy 
\def\aclpaperid{2314} 

\input{weld-defns}
\newcommand\sys{{\sc BoSsNet}}
\newcommand{\specialcell}[2][c]{%
\begin{tabular}[#1]{@{}c@{}}#2\end{tabular}}

\begin{document}
  
\title{Disentangling Language and Knowledge in Task-Oriented Dialogs}

\author{
Dinesh Raghu\thanks{\ \ D. Raghu is an employee at IBM Research. This work was carried out as part of PhD research at IIT Delhi.} $^{\, 1\, 2}$, 
Nikhil Gupta$^{1}$ and
Mausam$^1$ 
\\ 
$^1$ IIT Delhi, New Delhi, India\\
$^2$ IBM Research, New Delhi, India\\
{\em diraghu1@in.ibm.com,
nikhilgupta1997@gmail.com,
mausam@cse.iitd.ac.in}

}

\date{}

\maketitle
\begin{abstract}

The Knowledge Base (KB) used for real-world applications, such as booking a movie or restaurant reservation, keeps changing over time. End-to-end neural networks trained for these task-oriented dialogs are expected to be immune to any changes in the KB. However, existing approaches breakdown when asked to handle such changes. We propose an encoder-decoder architecture (\sys) with a novel Bag-of-Sequences (\textsc{BoSs}) memory, which facilitates the disentangled learning of the response's language model and its knowledge incorporation. Consequently, the KB can be modified with new knowledge without a drop in interpretability. We find that \sys\ outperforms state-of-the-art models, with considerable improvements (\textgreater10\%) on bAbI OOV test sets and other human-human datasets. We also systematically modify existing datasets to measure disentanglement and show \sys\ to be robust to KB modifications. 
%
\end{abstract}

\section{Introduction}

Task-oriented dialog agents converse with a user with the goal of accomplishing a specific task and often interact with a knowledge-base (KB). For example, a restaurant reservation agent \cite{hen2014word} will be grounded  to a KB that contains the names of restaurants, and their details. 

In real-world applications, the KB information could change over time. For example, (1) a KB associated with a movie ticket booking system gets updated every week based on new film releases, and (2) a restaurant reservation agent, trained with the knowledge of eateries in one city, may be deployed in other cities with an entirely different range of establishments. In such situations, the system should have the ability to conform to new-found knowledge unseen during its training. Ideally, the training algorithm must learn to disentangle the language model from the knowledge interface model. This separation will enable the system to generalize to KB modifications, without a loss in performance.

\begin{figure}[t]
\centering
\includegraphics[width=\columnwidth]{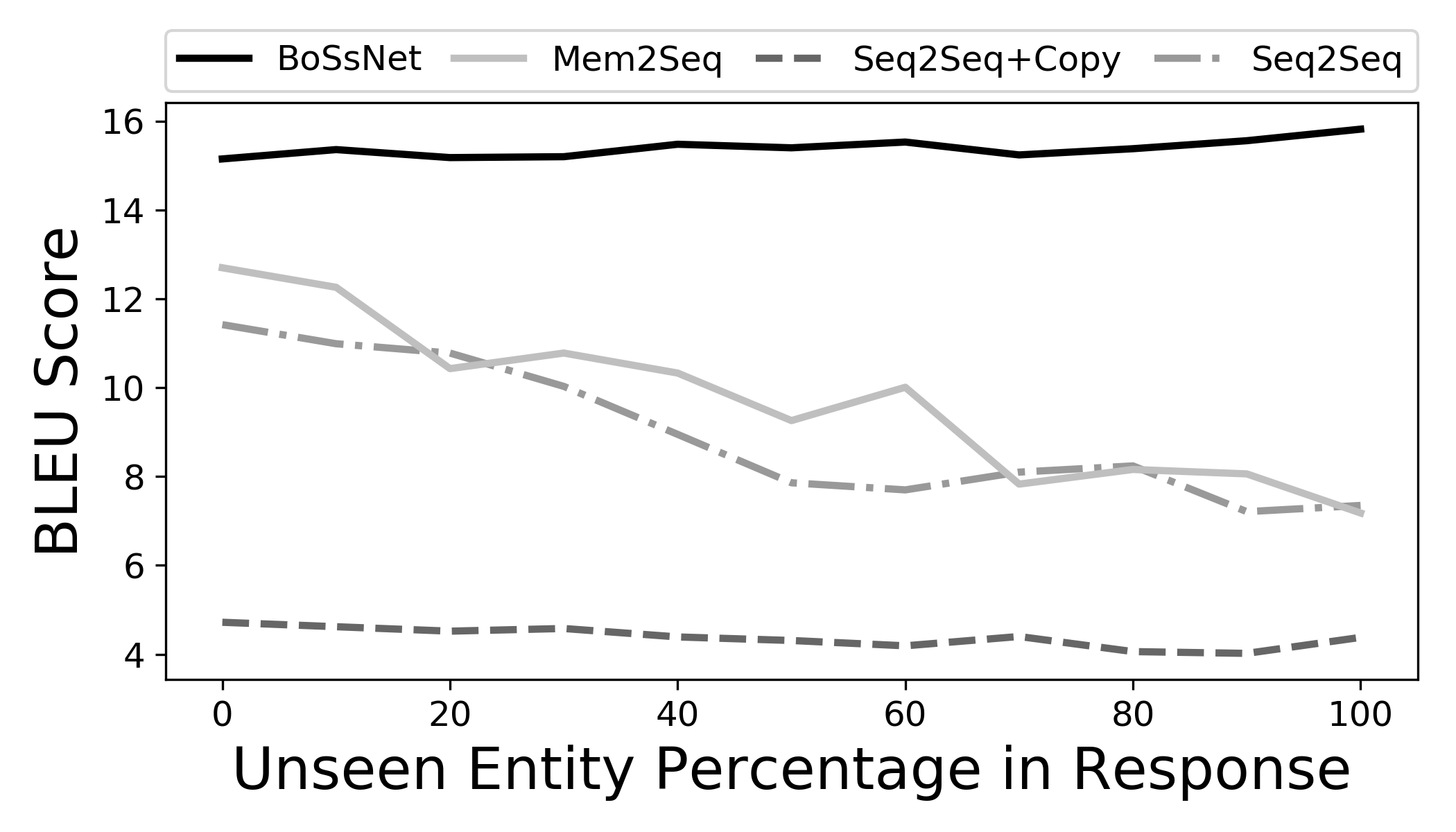}
\caption{Performance of various task-oriented dialog systems on the CamRest dataset as the percentage of unseen information in the KB changes.}
\label{fig:camrest}
\end{figure}

Moreover, for achieving good progress towards the user's task, the agent must also retain the ability to  draw inferences based on past utterances and the KB. Notably, we find that existing approaches either achieve this disentanglement or effective progress towards the task, but not both. 

For instance, Mem2Seq \cite{mem2seq} exhibits satisfactory performance when tested on the training KB. It represents the dialog history and the KB knowledge as a \emph{bag of words} in a flat memory arrangement. This enables Mem2Seq to revisit each word several times, as needed, obtaining good performance. But at the same time, flat memory prevents it from capturing any surrounding context -- this deteriorates its performance rapidly when the amount of new unseen information in the KB increases, as shown in Figure \ref{fig:camrest}. On the other hand, the performance of copy augmented sequence-to-sequence network (Seq2Seq+Copy) \cite{eric2017copy}, is robust to changes in the KB, but fails to achieve acceptable task-oriented performance. It captures context by representing the entire dialog history as one continuous \emph{sequence}.
However, it can be difficult for a sequence encoder to reason over long dialogs found in real-world datasets and its ability to learn the task gets hampered.

We propose \sys, a novel network that effectively disentangles the language and knowledge models, and also achieves state-of-the-art performance on three existing datasets. 

To achieve this, \sys\ makes two design choices. First, it encodes the conversational input as a {\em bag of sequences} (\textsc{BoSs}) memory, in which the input representation is built at two levels of abstraction. The \emph{higher level} flat memory encodes the KB tuples and utterances to facilitate effective inferencing over them. The \emph{lower level} encoding of each individual utterance and tuple is constructed via a sequence encoder (Bi-GRU). This enables the model to maintain the sequential context surrounding each token, aiding in better interpretation of unseen tokens at test time. Second, we augment the standard cross-entropy loss used in dialog systems with an additional loss term to encourage the model to only copy KB tokens in a response, instead of generating them via the language model. This combination of sequence encoding and additional loss (along with dropout) helps in effective disentangling between language and knowledge.

We perform evaluations over three datasets -- bAbI \cite{BordesW16}, CamRest \cite{wenEMNLP2016}, and Stanford Multi-Domain Dataset \cite{Ericsigdial}. Of these, the last two are real-world datasets. We find that \sys\ is competitive or significantly better on standard metrics in all datasets as compared to state-of-the-art baselines. We also introduce a {\em knowledge adaptability} (KA) evaluation, in which we systematically increase the percentage of previously unseen entities in the KB. We find that \sys\ is highly robust across all percentage levels. Finally, we also report a human-based evaluation and find that \sys\ responses are frequently rated higher than other baselines. 

Overall, our contributions are:

\begin{enumerate}
    \item We propose \sys, a novel architecture to disentangle the language model from knowledge incorporation in task-oriented dialogs.
    \item We introduce a {\em knowledge adaptability} evaluation to measure the ability of dialog systems to scale performance to unseen KB entities.
    \item Our experiments show that \sys\ is competitive or significantly better, measured via standard metrics, than the existing baselines on three datasets.
\end{enumerate}

We release our code and {\em knowledge adaptability} (KA) test sets for further use by the research community.\footnote{\url{ https://github.com/dair-iitd/BossNet}}

\section{The \sys\ Architecture}
\begin{figure}[t]
\centering
\includegraphics[scale=0.45]{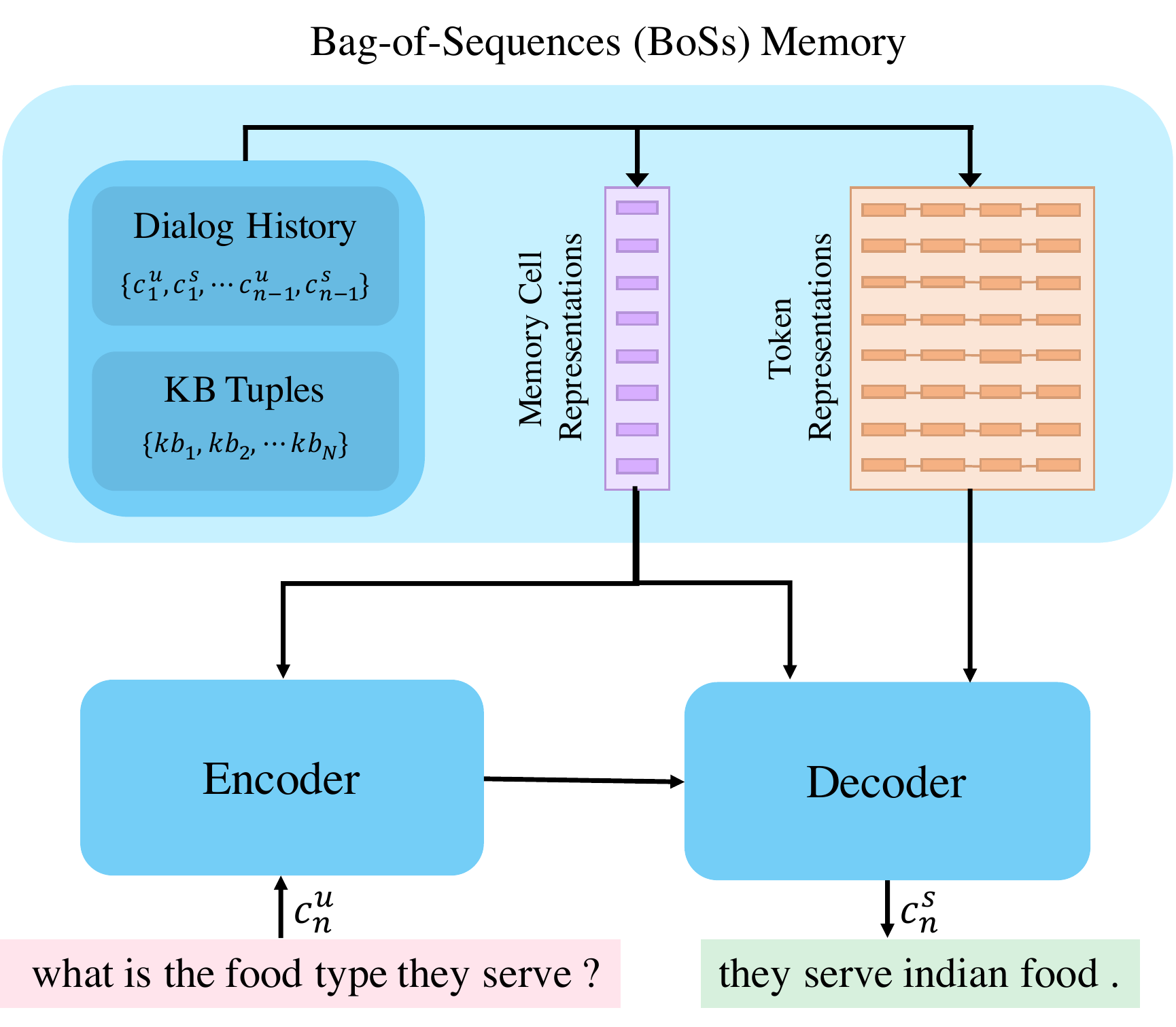}
\caption{The dialog history and KB tuples stored in the memory have memory cell representations and token representations. The encoder understands the last user utterance using only the memory cell representations. The decoder generates the next response using both representations.}
\label{fig:system}
\end{figure}

The proposed Bag-of-Sequences Memory Network has an encoder-decoder architecture that takes as input (1) dialog history, which includes a sequence of previous user utterances $\{c_1^u, \ldots, c_{n}^u\}$ and system responses $\{c_1^s, \ldots, c_{n-1}^s\}$, and (2) KB tuples $\{kb_1, \ldots, kb_{N}\}$. The network then generates the next system response $c_n^s=\langle y_1  y_2  \ldots  y_T \rangle$ word-by-word. The simplified architecture of \sys\ is shown in Figure \ref{fig:system}.

In this section, we first describe the {\sc BoSs} memory which contains the dialog history and KB tuples, followed by how the memory is consumed by the encoder and the decoder. We finally define the loss function, which, along with dropout, enables disentangled learning of language and knowledge.

\subsection{Bag-of-Sequences Memory} 
\label{sec:hmemory}
The memory $M$ contains the dialog history $\{c_1^u, c_1^s, \ldots, c_{n-1}^u, c_{n-1}^s\}$ and the KB tuples $\{kb_1, \ldots, kb_{N}\}$.  Each utterance in the dialog history and each KB tuple is placed in a memory cell. As utterances and tuples are inherently a sequence, we represent each  memory cell  $m_i$ as an ordered sequence of tokens $\langle w^1_i w^2_i \ldots w^{|m_i|}_i\rangle$. For an utterance, the word tokens are followed by a temporal indicator and a speaker indicator \{\$u, \$s\}. For example, \{\texttt{good, morning, \#1, \$s}\}\ indicates this was the first utterance by the system. For a KB tuple, the tokens are sequenced as \{\textit{subject, predicate, object}\} followed by temporal indicator and a kb indicator (\texttt{\$db}).

Token representation is generated using a bidirectional GRU. Let the outputs of the forward and backward GRUs for the token $w^j_i$ be denoted as $\overrightarrow{h^j_{i}}$ and $\overleftarrow{h^j_{i}}$ respectively. Then the token representation $\phi(w^j_i)$ is given by Eq. \ref{eqn:1}. Memory cell representation $\psi(m_i)$ is computed by concatenating the forward GRU output of its last token and the backward GRU output of its first token as in Eq. \ref{eqn:2}. 
\begin{eqnarray}
\phi(w^j_i)=[\overrightarrow{h^j_{i}};\overleftarrow{h_{i}^j}] \label{eqn:1} \\
\psi(m_i)=[\overrightarrow{h_{i}^{|m_i|}};\overleftarrow{h_{i}^1}] \label{eqn:2}
\end{eqnarray}


\subsection{The \sys\ Encoder}
\label{ssec:encoder}

The encoder used in \sys\ is similar to the multi-hop attention encoder with layer-wise weights proposed by \citeauthor{sukhbaatar2015end} (\citeyear{sukhbaatar2015end}). The encoder in \citeauthor{sukhbaatar2015end} (\citeyear{sukhbaatar2015end}) uses two different embedding matrices, whereas we use just one to reduce the number of parameters. The encoder considers the last user utterance as the query $q=\psi(c_n^u)$ and computes the reduced representation $q_r$ using the memory $M$ as follows:
\begin{eqnarray}
p_i &=& \text{softmax}(q^T \psi(m_i)) \\
o &=& W_r \sum\nolimits_i p_i \psi(m_i) \\
q_r &=& o + W_o  q
\end{eqnarray}

where $W_r, W_o \in \mathbb{R}^{d \times d}$ are learnable parameters. The hop step can be re-iterated, by assigning the output of the previous hop as the new input query, i.e., setting $q=q_r$. The output of the encoder after $K$ hops, $q_r^k$, is assigned as the initial state of the \sys\ decoder.

\subsection{The \sys\ Decoder}
\label{ssec:decoder}

\sys\ models a copy-augmented sequence decoder, which generates the response one word at a time. At any decode time step $t$, the decoder can either \emph{generate} a word from the decode vocabulary or \emph{copy} a word from the memory. Consequently, the decoder computes: (1) generate distribution $P_g(y_t)$ over the decode vocabulary, and (2) copy distribution $P_c(y_t)$ over words in the memory. 

The generate distribution is computed using a standard sequence decoder \cite{sutskever2014sequence} by attending \cite{luong2015effective} over the memory cell representations $\psi$. The copy distribution is generated by using a \textit{two-level attention}. Given the decoder state $s_t$, it first computes attention $\alpha_t$ over the memory cells. Then it computes attention over the tokens in each memory cell $m_i$. Finally it multiplies both these attentions to compute $P_c(y_t)$ as follows: 
\begin{gather}
\alpha_i^t = \text{softmax}(s_t \psi(m_i)) \\
e_{ij}^t = s_t \phi(w_i^j) \\
\beta^{t}_{ij} = \alpha^t_i * \frac{\exp({e_{ij}^t})}{\sum\nolimits_{k}\exp({e_{ik}^t})} \\
P_c(y_t=w)=\sum_{ij:w_i^j=w} \beta_{ij}^{t}
\end{gather}


The copy and generate distributions are combined using a soft gate $g_s^t \in [0,1]$ as in \citeauthor{see2017get} (\citeyear{see2017get}). $g_s^t$ is a function of the decoder state at time $t$ and the word decoded in the previous time step.

\subsection{Loss}
The decoder is trained using cross-entropy loss. The loss per response is defined as:
\begin{equation}
\mathcal{L}_{ce} = - \sum_{t=1}^{T} \textup{log} \Big( g_s^{t}P_g(y_t) + (1-g_s^{t})P_c(y_t) \Big) 
\end{equation}
where $T$ is the number of words in the sequence to be generated and $y_t$ is the word to be generated at time step $t$. The decision to generate or copy is learnt implicitly by the network. However, to attain perfect disentanglement, the KB words should be copied, while the language should be generated. In other words, any word in the response that is present in the {\sc BoSs} KB memory should have a low $g_s$. To obtain this behavior, we define a disentangle label $D_{l}$ for each word in the response. This label is set to $1$ if the word is present in the {\sc BoSs} KB memory and $0$ otherwise.
We define a disentangle loss as follows:
\begin{equation}
\mathcal{L}_{d} = - \sum_{t=1}^{T}  g_s^{t}\textup{log}D^{t}_{l} + (1-g_s^{t})\textup{log}(1-D^{t}_{l})
\end{equation}

We randomly drop some words with disentangle label set to $1$. This \textit{Disentangle Label Dropout (DLD)} works in tandem with the disentangle loss and {\sc BoSs} memory -- it encourages the model to copy KB words whenever possible, based on their surrounding words.  The overall loss is given as:
\begin{equation}
\mathcal{L} = \mathcal{L}_{ce} + \gamma \mathcal{L}_{d}
\label{eqn:loss}
\end{equation}

The relative weight of $\mathcal{L}_{d}$ in the overall loss is controlled using a hyper-parameter ($\gamma$). The dropout rate is also a hyper-parameter.

\section{Experimental Setup}


We perform experiments on three task-oriented dialog datasets: bAbI Dialog \cite{BordesW16}, CamRest \cite{wenEMNLP2016}, and Stanford Multi-Domain Dataset \cite{Ericsigdial}.

\noindent 
\textbf{bAbI Dialog} consists of synthetically generated dialogs with the goal of restaurant reservation. The dataset consists of five different tasks, all grounded to a KB. This KB is split into two mutually exclusive halves. One half is used to generate the train, validation, and test sets, while the other half is used to create a second test set called the OOV test set. 


\noindent 
\textbf{CamRest} is a human-human dialog dataset, collected using the Wiz-of-Oz framework, also aimed at restaurant reservation. It is typically used to evaluate traditional slot filling systems. In order to make it suitable for end-to-end learning, we stripped the handcrafted state representations and annotations in each dialog, and divided the 676 available dialogs into train, validation, and test sets (406, 135, and 135 dialogs, respectively).

\noindent
\textbf{Stanford Multi-Domain Dataset (SMD)} is another human-human dialog dataset collected using the Wiz-of-Oz framework. Each conversation is between a driver and an in-car assistant. The other datasets consist of dialogs from just one domain (restaurant reservation), whereas SMD consists of dialogs from multiple domains (calendar scheduling, weather information retrieval, and navigation).

\subsection{Knowledge Adaptability (KA) Test Sets}
Each bAbI dialog task has an additional OOV test set, which helps to evaluate a model's robustness to change in information in the KB. A model that perfectly disentangles language and knowledge should have no drop in accuracy on the OOV test set when compared to the non-OOV test set. To measure the degree of disentanglement in a model, we generated 10 additional test sets for each real-world corpus by varying the percentage (in multiples of 10) of unseen entities in the KB. We systematically picked random KB entities and replaced all their occurrences in the dialog with new entity names. We will refer to these generated dialogs as the \emph{Knowledge Adaptability} (KA) test sets.

\subsection{Baselines}
We compare \sys\ against several existing end-to-end task-oriented dialog systems. These include retrieval models, such as the query reduction network (QRN) \cite{seo2016query}, memory network (MN) \cite{BordesW16}, and gated memory network (GMN) \cite{liu2017gated}. 
We also compare against generative models such as a sequence-to-sequence model (Seq2Seq), a copy augmented Seq2Seq (Seq2Seq+Copy) \cite{ptr-unk}, and Mem2Seq \cite{mem2seq}.\footnote{We thank the authors for releasing a working code at \url{https://github.com/HLTCHKUST/Mem2Seq}} For fairness across models, we do not compare against key-value retrieval networks \cite{Ericsigdial} as they simplify the dataset by canonicalizing all KB words in dialogs.

We noticed that the reported results in the Mem2Seq paper are not directly comparable, as they pre-processed\footnote{Mem2Seq used the following pre-processing on the data: 1) The subject (restaurant name) and object (rating) positions of the rating KB tuples in bAbI dialogs are flipped 2) An extra fact was added to the navigation tasks in SMD which included all the properties (distance, address, etc.) combined together  as the subject and \textit{poi} as the object. See Appendix.} training data in SMD and bAbI datasets. For fair comparisons, we re-run Mem2Seq on the original training datasets. For completeness we mention their reported results (with pre-processing) as Mem2Seq*.

\subsection{Evaluation Metrics}
We evaluate \sys\ and other models based on their ability to generate valid responses. The per-response accuracy \cite{BordesW16} is the percentage of generated responses that exactly match their respective gold response. The per-dialog accuracy is the percentage of dialogs with all correctly generated responses. These accuracy metrics are a good measure for evaluating datasets with boilerplate responses such as bAbI. 

To quantify performance on other datasets, we use BLEU \cite{papineni2002bleu} and Entity F1 \cite{eric2017copy} scores. BLEU measures the overlap of n-grams between the generated response and its gold response and has become a popular measure to compare task-oriented dialog systems. Entity F1 is computed by micro-F1 over KB entities in the entire set of gold responses. 

\subsection{Human Evaluation}
We use two human evaluation experiments to compare (1) the \emph{usefulness} of a generated response with respect to solving the given task, and (2) the \emph{grammatical correctness} and \emph{fluency} of the responses on a 0--3 scale. We obtain human annotations by creating Human Intelligence Tasks (HITs) on Amazon Mechanical Turk (AMT). For each test condition (percentage of unseen entities), we sampled 50 dialogs from Camrest and SMD each, and two AMT workers labeled each system response for both experiments, resulting in 200 labels per condition per dataset per system. We evaluate four systems in this study, leading to a total of 1600 labels per condition. The detailed setup is given in the Appendix. 

\subsection{Training}
We train \sys\ using an Adam optimizer \cite{kingma2014adam} and apply gradient clipping with a clip-value of 40. We identify hyper-parameters based on the evaluation of the held-out validation sets. We sample word embedding, hidden layer, and cell sizes from \{64, 128, 256\} and learning rates from \{10$^{-3}$, 5$\times$10$^{-4}$, 10$^{-4}$\}. The hyper-parameter $\gamma$ in the loss function is chosen between [0-1.5]. The Disentangle Label Dropout rate is sampled from \{0.1, 0.2\}. The number of hops for multi-hop attention in the encoder is sampled from \{1, 3, 6\}. The best hyper-parameter setting for each dataset is reported in the Appendix.

\begin{table*}[ht]
\centering
\footnotesize
\begin{tabular}{c|ccc|cccc|c}
\toprule
& \multicolumn{3}{c|}{\textbf{Retrieval Models}} & \multicolumn{4}{c}{\textbf{Generative Models}}  \\
\cmidrule{2-9}
\textbf{Task} & \textbf{QRN} & \textbf{MN} &  \textbf{GMN} & \textbf{Mem2Seq*} & \textbf{Seq2Seq} & \textbf{Seq2Seq+Copy} & \textbf{Mem2Seq} & \textbf{ \sys\ }  \\
\midrule
T1 & 99.9 (-) & 99.6 (99.6) & 100 (100) & 100 (100) & 100 (100) & 100 (100) & 100 (100) & 100 (100) \\
T2 & 99.5 (-) & 100 (100) & 100 (100)& 100 (100) &  100 (100) &  100 (100) & 100 (100) & 100 (100) \\
T3 & 74.8 (-) & 74.9 (2.0) & 74.9 (0)& 94.7 (62.1) & 74.8 (0) & 85.1 (19.0)& 74.9 (0) & \textbf{95.2 (63.8)}    \\
T4 & 57.2 (-) & 59.5 (3.0) & 57.2 (0)& 100 (100) & 57.2 (0) & 100 (100) & 100 (100)  & 100 (100) \\
T5 & \textbf{99.6 (-)} & 96.1 (49.4) & 96.3 (52.5) & 97.9 (69.6) & 97.2 (64.4) & 96 (49.1)  & 97.7(66.3) & 97.3 (65.6) \\
\midrule
T1-OOV & 83.1 (-) & 72.3 (0) & 82.4 (0) & 94.0 (62.2) & 81.7 (0) & 92.5 (54.7)  & 94.0 (62.2) & \textbf{100 (100)}\\
T2-OOV & 78.9 (-) & 78.9 (0) & 78.9 (0)& 86.5 (12.4)  & 78.9 (0) & 83.2 (0) & 86.5 (12.4) & \textbf{100 (100)}\\
T3-OOV & 75.2 (-) & 74.4 (0) & 75.3 (0)& 90.3 (38.7) & 75.3 (0) & 82.9 (0) & 75.2 (0) &  \textbf{95.7 (66.6)}   \\
T4-OOV & 56.9 (-) & 57.6 (0) & 57.0 (0)& 100 (100) & 57.0 (0) & 100 (100) & 100 (100) & 100 (100) \\
T5-OOV & 67.8 (-) & 65.5 (0) & 66.7 (0)& 84.5 (2.3) & 67.4 (0) & 73.6 (0) & 75.6 (0) & \textbf{91.7 (18.5)}   \\
\bottomrule
\end{tabular}
\caption{Per-response and per-dialog accuracies (in brackets) on bAbI dialog tasks of \sys\ and baselines}. 
\label{tab:babi}
\end{table*}

\section{Experimental Results}
\label{sec:experiments}

Our experiments evaluate three research questions. 
\begin{compactenum}
    \item \emph{Performance Study}: How well is \sys\ able to perform the tasks of our three datasets as compared to the baseline models? 
    \item \emph{Disentanglement Study}: How robust are the models in generalizing on the KA test sets? 
    \item \emph{Ablation Study}: What is the performance gain from each novel feature in \sys? 
\end{compactenum}

\subsection{Performance Study}
\label{sec:expt2}

Table \ref{tab:babi} reports the per-response and per-dialog (in parentheses) accuracies on the bAbI dialog tasks.
The multi-hop retrieval-based models such as QRN, MN and GMN perform well on the non-OOV test sets for tasks 1, 2, and 5, but fail to exhibit similar performance on the corresponding OOV test sets. This result is expected as these models are trained to retrieve from a pre-defined set of responses. Their poor non-OOV performance on tasks 3 and 4 is attributed to an error in the bAbI dataset construction, due to which, the non-OOV and OOV test conditions are the same for these tasks (see Appendix).

A simple generative model (Seq2Seq) achieves accuracies comparable to the multi-hop retrieval models. Enabling it with the ability to copy from the context (Seq2Seq+Copy) shows a considerable increase in performance, especially on the OOV test sets (and non-OOV tests for tasks 3 and 4).

The strong performance of simple sequence encoders when compared with multi-hop encoders (in retrieval models) raises a question about the value of multi-hop inference. Mem2Seq answers this question, by obtaining improvements in several tasks,  specifically on their OOV test sets. This clearly shows that multi-hop inference and the copy mechanism are essentials for task-oriented dialogs.

Despite gains from the Mem2Seq model, the performance difference between the non-OOV and OOV test sets remains large. \sys\ succeeds to bridge this gap with its ability to better interpret unseen words, using their surrounding context. It obtains significant improvements on average of about 34\% per-dialog accuracy and 10\% per-response accuracy for the bAbI OOV test sets.



\begin{table}[t]
\centering
\footnotesize
 \begin{tabular}{l|cc|cc}
\toprule
& \multicolumn{2}{c|}{\textbf{CamRest}} & \multicolumn{2}{c}{\textbf{SMD}}  \\ \cmidrule{2-5}
& \textbf{BLEU} & \textbf{Ent. F1} & \textbf{BLEU} & \textbf{Ent. F1} \\
\midrule
Mem2Seq* & 12.7 & 39 & 12.6 & 33.4  \\
\midrule
Seq2Seq & 11.4 & 40.6 & 8.7 & 34.9  \\
Seq2Seq+Copy & 4.7 & 32.2 & 3.23 & 16.9  \\
Mem2Seq & 12.7 & 39 & \textbf{10.3} & 31.8 \\ 
\midrule
\sys\ & \textbf{15.2} & \textbf{43.1} & 8.3 & \textbf{35.9} \\
\bottomrule
\end{tabular}
\caption{Performance of \sys\ and baselines on the CamRest and SMD datasets}
\label{tab:smd}
\end{table}

In Table \ref{tab:smd}, we report results on the real-world datasets. \sys\ greatly outperforms other models in both Entity F1 metric and BLEU scores on CamRest. On SMD, \sys\ achieves the best only in Entity F1. On further analysis of the generated responses we observe that \sys\ responses often convey the necessary entity information from the KB. However, they consist of meaningful phrases with little lexical overlap with the gold response, reducing the BLEU scores. We investigate this further in our human evaluation.

\noindent \textbf{Human Evaluation:}
We summarize the human evaluation results for real-world datasets in Table \ref{tab:amt_perf}. \sys\ shows the best performance on Camrest, and is judged useful 77 times out of 100. Also, it has the highest average grammatical correctness score of 2.28 (very close to Seq2Seq and Mem2Seq). \sys\ performs on par with Mem2Seq and Seq2Seq in its ability to relay appropriate information to solve SMD dialog tasks, and has a slightly higher grammaticality score.

\begin{table}[t]
\centering
\footnotesize
 \begin{tabular}{l|cc|cc}
\toprule
& \multicolumn{2}{c|}{\textbf{CamRest}} & \multicolumn{2}{c}{\textbf{SMD}}  \\ \cmidrule{2-5}
& \textbf{Info} & \textbf{Grammar} & \textbf{Info} & \textbf{Grammar} \\
\midrule
Seq2Seq & 46 & 2.24 & 35 &  2.38 \\
Seq2Seq+Copy & 27 & 1.1 & 21 &  1.04 \\
Mem2Seq & 51 & 2.2 & \textbf{38} &  2.0 \\
\midrule
\sys\ & \textbf{77} & \textbf{2.28} & 36 &  \textbf{2.5} \\

\bottomrule
\end{tabular}
\caption{AMT Evaluations on CamRest and SMD} 
\label{tab:amt_perf}
\end{table}

\subsection{Disentanglement Study}
\label{sec:expt1}
We use our generated knowledge adaptability (KA) test sets to measure the robustness of \sys\ and the other baselines to changes in the KB. We perform this experiment on 4 different tasks, namely bAbI tasks 1 and 5, CamRest, and SMD.

\begin{figure*}
\centering
\begin{minipage}[b]{.475\textwidth}
\includegraphics[width=\textwidth, height=4.7cm]{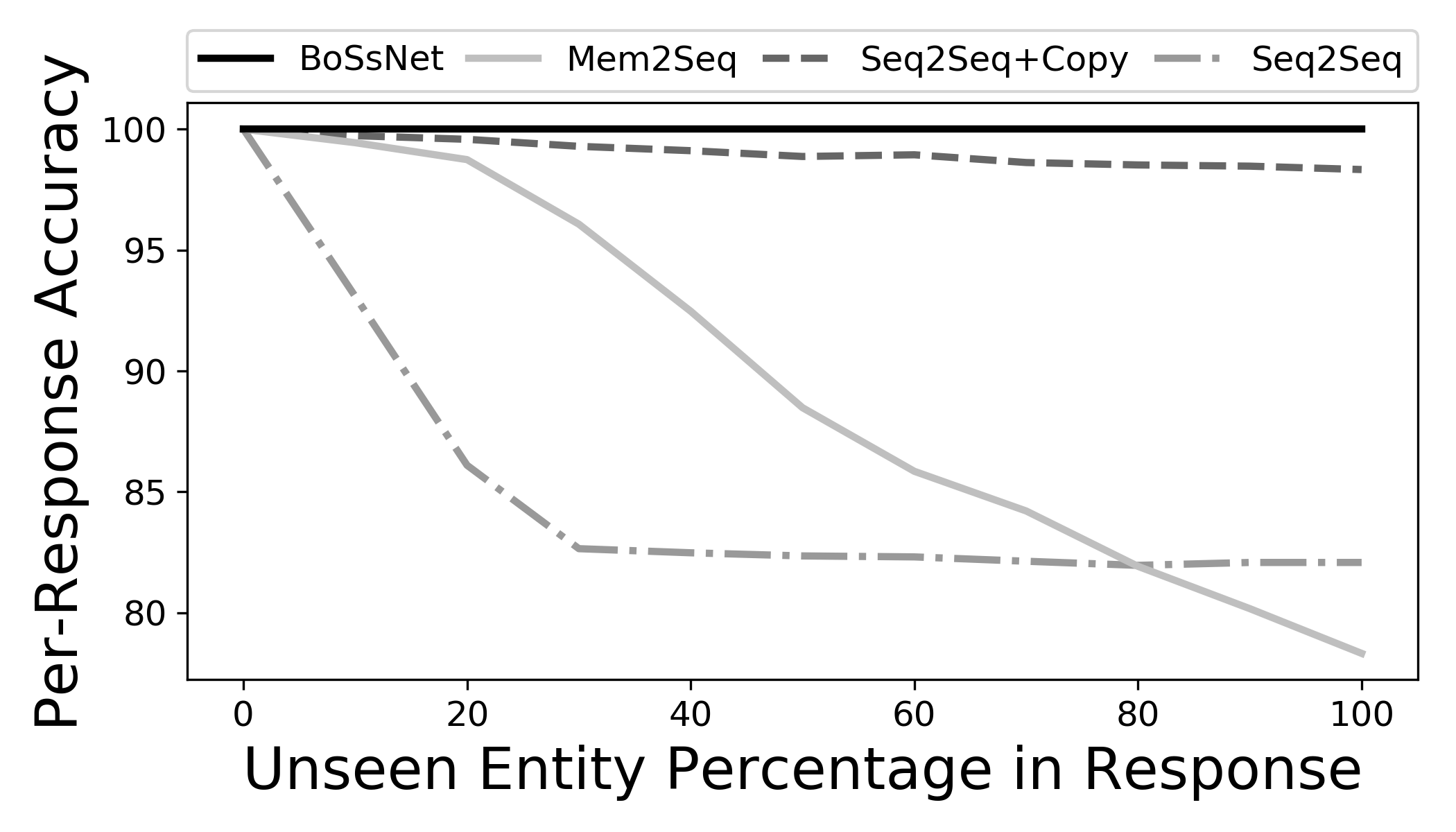}
\caption{bAbI Task 1: Per-response accuracy comparison on KA sets}\label{label-a}
\end{minipage}\qquad
\begin{minipage}[b]{.475\textwidth}
 \includegraphics[width=\textwidth, height=4.7cm]{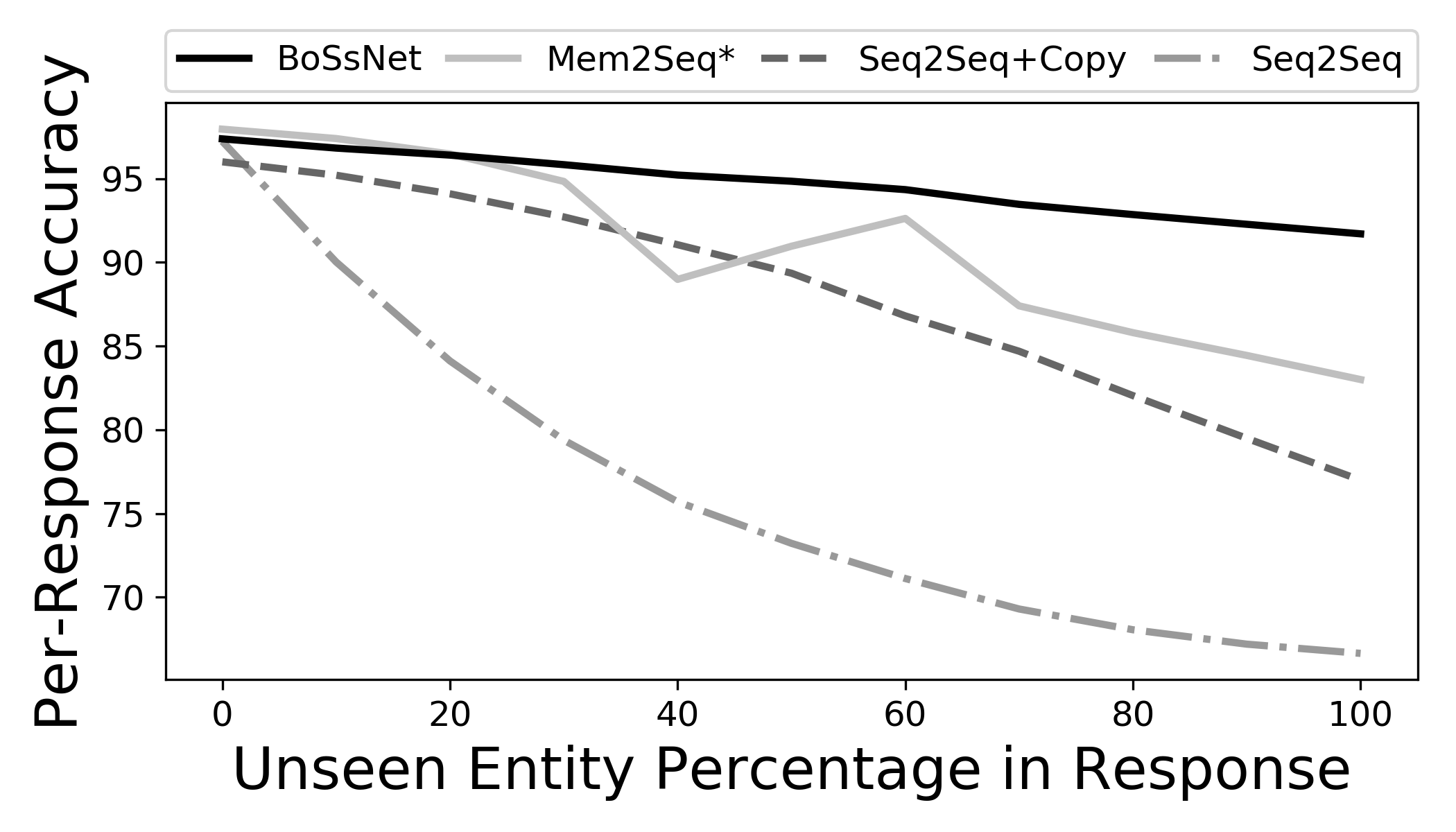}
        \caption{bAbI Task 5: Per-response accuracy comparison on KA sets}\label{label-b}
\end{minipage}
\end{figure*}

Figures \ref{label-a} and \ref{label-b} show the per-response accuracies of the two bAbI dialog tasks plotted against the percentage of unseen entities in KA sets. From Figure \ref{label-a} we observe that \sys\ remains immune to any variablity in the KB content, whereas the performance of Mem2Seq and Seq2Seq models drops drastically due to their inability to capture semantic representations of the injected KB entities. We see a similar trend in Figure \ref{label-b}, but here all the models show a drop in performance, with \sys\ appearing the most steady. We explain this trend using the example dialog in Table \ref{tab:t5_dis}. In the current dialog context, the system is required to provide the address of the selected restaurant, but since more than one restaurant in the KB is unseen, it becomes ambiguous for the network to identify the correct restaurant and infer its address. In the end, the system is forced to pick a random address -- the probability of which being correct decreases as more restaurants become unseen.

\begin{table}[!t]
\centering
\scriptsize
\begin{tabular}{c|l}
\toprule
\multicolumn{2}{c}{\textbf{KB (restaurant|address)}} \\
\multicolumn{2}{c}{\textit{r\_bangkok\_overpriced\_thai\_8}|\textit{r\_bangkok\_overpriced\_thai\_8\_addr}}\\
\multicolumn{2}{c}{\textit{r\_bangkok\_overpriced\_thai\_7}|\textit{r\_bangkok\_overpriced\_thai\_7\_addr}}\\
\multicolumn{2}{c}{\textit{r\_bangkok\_overpriced\_thai\_4}|\textit{r\_bangkok\_overpriced\_thai\_4\_addr}}\\
\multicolumn{2}{c}{\textit{r\_bangkok\_overpriced\_thai\_2}|\textit{r\_bangkok\_overpriced\_thai\_2\_addr}}\\
\midrule
\midrule
\textbf{usr-1} & may i have a table in an \textit{overpriced} price range for \\
& \textit{nine} people with \textit{thai} food in \textit{bangkok} ? \\
\textbf{sys-1} & what do you think of : \textit{r\_bangkok\_overpriced\_thai\_8} ? \\
\textbf{usr-2} & can you provide the address ? \\
\midrule
\textbf{Gold} & here it is \textit{r\_bangkok\_overpriced\_thai\_8\_addr}
 \\
\midrule
\midrule
\textbf{Seq2Seq+Copy} & here it is \textit{r\_bangkok\_overpriced\_thai\_4\_addr}
 \\
\midrule
\textbf{Seq2Seq} & here it is \textit{r\_london\_moderate\_spanish\_6\_addr} \\

\midrule
\textbf{Mem2Seq} & here it is \textit{r\_bangkok\_overpriced\_thai\_4\_addr} \\
\midrule
\textbf{\sys\ } & here it is \textit{r\_bangkok\_overpriced\_thai\_4\_addr} \\
\bottomrule
\end{tabular}
\caption{Example from bAbI Task 5 KA test set with 100\% OOV entities. Identifying the address of an unseen restaurant is challenging for all models.}
\label{tab:t5_dis}
\end{table}

\begin{figure*}
\centering
\begin{minipage}[b]{.475\textwidth}
\includegraphics[width=\textwidth, height=4.7cm]{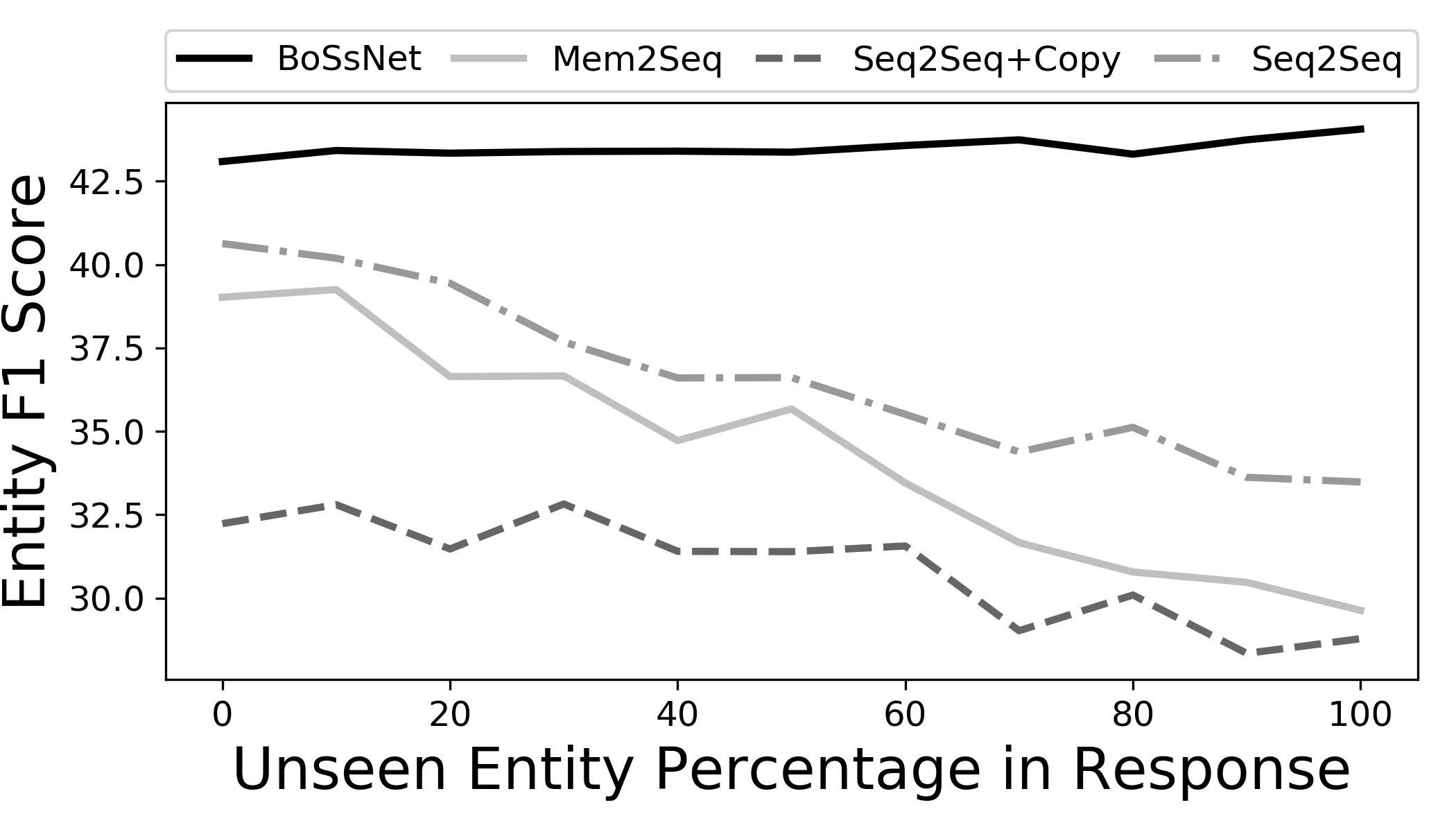}
\caption{CamRest: Entity F1 comparison on KA sets}\label{label-c}
\end{minipage}\qquad
\begin{minipage}[b]{.475\textwidth}
 \includegraphics[width=\textwidth, height=4.7cm]{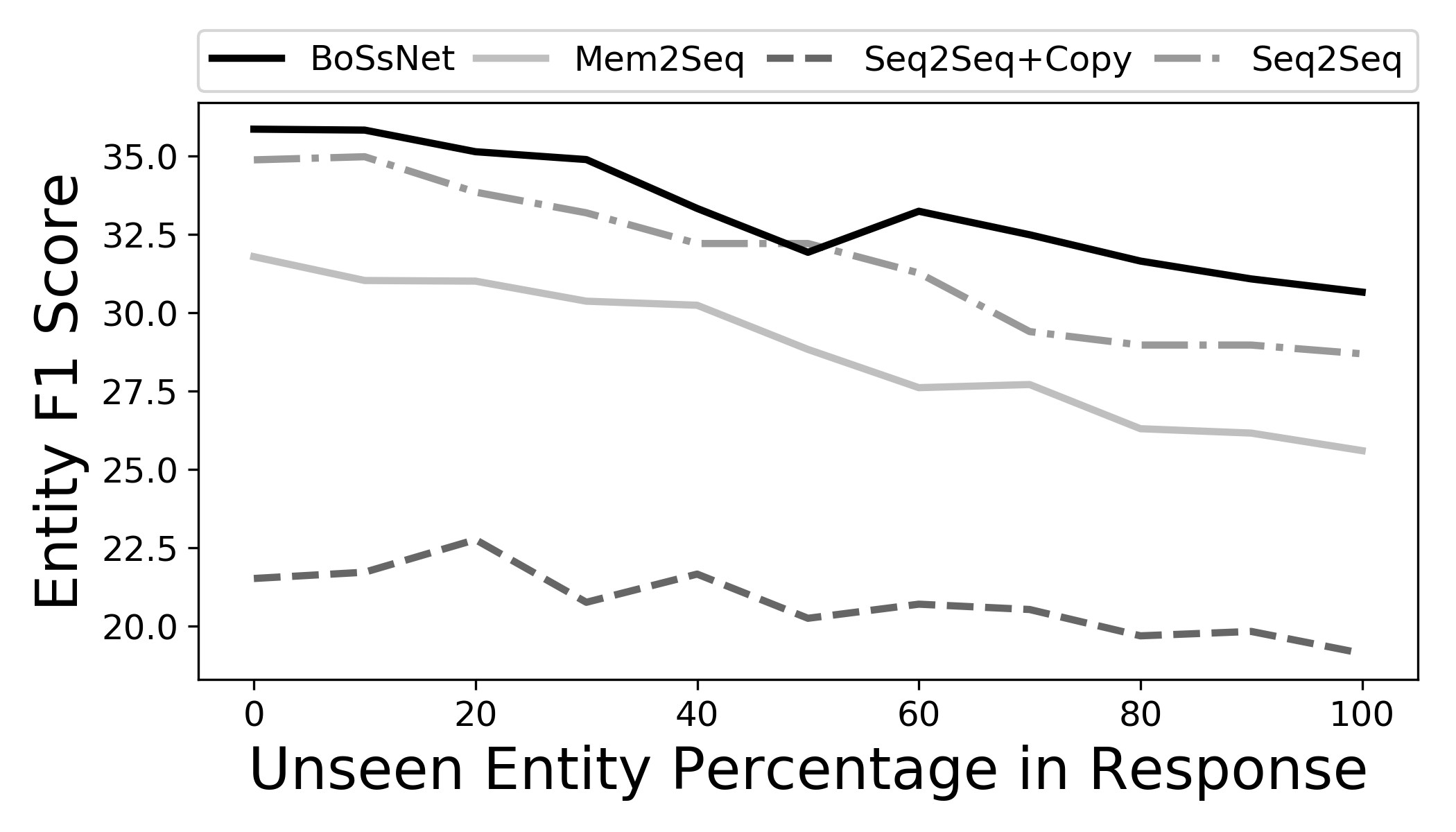}
\caption{SMD: Entity F1 comparison on KA sets}\label{label-d}
\end{minipage}
\end{figure*}

The performance on the CamRest KA test sets is illustrated in Figures \ref{fig:camrest} and \ref{label-c}. \sys\ has the best performance with even a slight increase in both BLEU and Entity F1 metrics as more OOV content is injected in the dialog, probably because it is clear that it needs to copy when processing unseen entities.
Seq2Seq+Copy is unable to perform well in CamRest as the length of the input (dialog history + KB tuples) is long and the size of the training set is also small. We believe that Seq2Seq+Copy works best in an environment with an abundance of short dialog training data (e.g., bAbI task 1 in Figure \ref{label-a}).

SMD consists of dialogs with a large KB and a highly varying response pattern. This makes it very difficult to learn the language model -- reflected in the low BLEU scores for all the systems. \sys\ still provides the best F1 entity score due to its ability to inference efficiently on the large KB (Figure \ref{label-d}). Mem2Seq shows the best BLEU score performance on the original test set, but its performance drop of 42.5\%, from 10.3 at 0\% unseen to 5.93 at 100\% unseen, is a lot heavier than that of \sys\ which only drops 7.6\% -- 8.27 at 0\% unseen to 7.64 at 100\% unseen.

\noindent \textbf{Human Evaluation:}
We summarize the human evaluation results for real-world datasets on the 50\% unseen KA test set in Table \ref{tab:amt_dis}. \sys\ again outperforms the baselines and is labeled \emph{successful} twice more often than the next best model on both Camrest and SMD. Seq2Seq appears to produce better sentence structures on the SMD dataset, primarily because it does not attempt to learn inference on the KB, allowing it to solely focus on learning the language model better. 

\begin{table}[t]
\centering
\footnotesize
 \begin{tabular}{l|cc|cc}
\toprule
& \multicolumn{2}{c|}{\textbf{CamRest}} & \multicolumn{2}{c}{\textbf{SMD}}  \\ \cmidrule{2-5}
& \textbf{Info} & \textbf{Grammar} & \textbf{Info} & \textbf{Grammar} \\
\midrule
Seq2Seq & 26 & 2.28 & 22 & \textbf{2.44} \\
Seq2Seq+Copy & 22 & 1.22 & 16 & 1.04 \\
Mem2Seq & 35 & 2.06 & 26 & 1.9 \\
\midrule
\sys\ & \textbf{80} & \textbf{2.44} & \textbf{51} &  2.28 \\

\bottomrule
\end{tabular}
\caption{AMT Evaluations on CamRest and SMD (50\% unseen) KA datasets} 
\label{tab:amt_dis}
\end{table}





\subsection{Ablation Study}
\label{sec:expt3}

\begin{table*}[!ht]
\centering
\footnotesize
\begin{tabular}{l|ccccc|ccccc|cc}
\toprule
   & \multicolumn{5}{c|}{\textbf{bAbI Dialog Tasks}} & \multicolumn{5}{c|}{\textbf{bAbI Dialog Tasks (OOV)}}  & \multicolumn{2}{c}{\textbf{CamRest}} \\ \cmidrule{2-6} \cmidrule{7-11} \cmidrule{12-13}
    & T1  & T2  & T3   & T4   & T5   & T1 & T2 & T3 & T4 & T5 & BLEU        & Ent. F1       \\ \midrule
\sys\ w/o {\sc BoSs} Memory & 100 & 100 & 74.9 & 57.2 & 95.6 & 93.5   & 78.9   & 74.9   & 57     & 81.4   & 10.13        & 29          \\
\sys\ w/o $\mathcal{L}_{d}$          & 100 & 100 & 91.7 & 100  & 94.3 & 83.2   & 78.9   & 92.7   & 100    & 66.7   & 15.5        & 40.1          \\
\sys\  w/o DLD     & 100 & 100 & 93.4 & 100  & 95.3 & 79.2   & 84.6   & 90.7   & 100    & 78.1   & 12.4        & 40.45         \\ \midrule
\sys\                 & 100 & 100 & 95.2 & 100  & 97.3 & 100    & 100    & 95.7   & 100    & 91.7   & 15.2        & 43.1    
\\ \bottomrule
\end{tabular}
\caption{Ablation study: impact of each model element on \sys\ }
\label{tab:ablation}
\end{table*}

We assess the value of each model element, by removing it from \sys. Table \ref{tab:ablation} reports the per-response accuracy scores for various configurations of \sys\ on bAbI dialog tasks. It also reports the BLEU and entity F1 metric of various configurations on CamRest.

\noindent \textbf{Without BoSs Memory:} 
This configuration uses the Bag-of-Bags (BoB) Memory rather than {\sc BoSs} memory. The BoB memory is a simplified representation, similar to the one in the original Memory Networks. Here the token representation is the vector embedding of the token with no influence from the surrounding words and the memory cell representation is the sum of all its token embeddings. As a result, each word $w$ representation is influenced equally by all words in a memory cell, irrespective of its distance from $w$. This makes capturing context in the immediate neighbourhood harder. Inability to capture the correct context prevents the configuartion from generalizing to OOV test sets.

\noindent \textbf{Without Disentangled Loss:} Disentangled Loss ($\mathcal{L}_{d}$) plays an important role in enforcing that KB words be copied and other language be generated. By removing this loss component, 
it achieves better BLEU score in CamRest, but with a drop in Entity F1. Without the disentangled loss, the model sometimes learns to generate KB words. This severely affects OOV performance. As described earlier, an error in bAbI dataset construction tasks 3 and 4 effectively injects the validation set with a lot of OOVs. This anomaly in conjunction with the dropout (DLD), helps the configuration in achieving an acceptable performance for those tasks.

\noindent \textbf{Without Disentangled Label Dropout:} 
\sys\ learns to generate language and copy KB words. Without DLD, the model learns to memorize words to be copied rather than learning the context under which a word should be copied. Hence, the performance on OOV test sets is much inferior compared to the non-OOV setting.

Overall, we notice that combining all three model elements is necessary in obtaining the best performance across all tasks.


\subsection{Qualitative Evaluation}
We qualitatively compare the performance of \sys\ with other baselines using examples.

Table \ref{tab:camrest-qualeval}, demonstrates the ability of \sys\ to copy entities (restaurant name and address) in its response. The other baselines either generate unwanted or irrelevant entities in their response, or fail to copy altogether. \sys\ also best captures the language model effectively with a slight paraphrasing of the gold response.

Table \ref{tab:task5-qualeval} contains only unseen entities. This example highlights the shortcomings of the Seq2Seq model as it ends up predicting a restaurant encountered during training. Mem2Seq copies a restaurant name without learning to sort the restaurants based on rating. \sys, with its efficient memory addressing, is seen to be able to solve both issues.

\begin{table*}[h]
\centering
\small
\begin{tabular}{c|p{13cm}}
\toprule
\multicolumn{2}{c}{\textbf{KB (restaurant|cuisine|address|phone)}} \\
\multicolumn{2}{c}{\textit{pizza\_hut\_fen\_ditton}|\textit{italian}|\textit{cambridge\_retail\_park\_newmarket\_road\_fen\_ditton}|\textit{01223\_323737}}\\
\midrule
\midrule
\textbf{usr-1} & may i have information for an \textit{italian} restaurant in the \textit{east} part of town ? \\
\textbf{sys-1} & yes sure . there is only on \textit{italian} restaurant called \textit{pizza\_hut\_fen\_ditton} in the \textit{east} . what else do you want to know ? \\
\textbf{usr-2} & what is their address please ? \\
\midrule
\textbf{Gold} & the address for \textit{pizza\_hut\_fen\_ditton} is \textit{cambridge\_retail\_park\_newmarket\_road\_fen\_ditton} . can i help you with anything else ?
 \\
\midrule
\midrule
\specialcell[t]{\textbf{Seq2Seq+Copy}} & their address is at
 \\
\midrule
\textbf{Seq2Seq} & \textit{curry\_prince} is at \textit{451\_newmarket\_road\_fen\_ditton} \\

\midrule
\textbf{Mem2Seq} & \textit{pizza\_hut\_fen\_ditton} is located at \textit{7\_barnwell\_road\_fen\_ditton} is \textit{7\_barnwell\_road\_fen\_ditton} and the phone number is \textit{01223\_244955}
 \\
\midrule
\textbf{\sys\ } & \textit{pizza\_hut\_fen\_ditton} is located at \textit{cambridge\_retail\_park\_newmarket\_road\_fen\_ditton} is there anything else i can help you with \\
\bottomrule
\end{tabular}
\caption{An example of responses generated by \sys\ and baselines on the CamRest test set. Thia example has no unseen entities.}
\label{tab:camrest-qualeval}
\end{table*}

\begin{table*}
\centering
\small
\begin{tabular}{c|l}
\toprule
\multicolumn{2}{c}{\textbf{KB (restaurant|cuisine|price\_range|location|rating)}} \\
\multicolumn{2}{c}{\textit{resto\_tokyo\_affordable\_vietnamese\_8stars}|\textit{vietnamese}|\textit{affordable}|\textit{tokyo}|\textit{8}}\\
\multicolumn{2}{c}{\textit{resto\_tokyo\_affordable\_vietnamese\_7stars}|\textit{vietnamese}|\textit{affordable}|\textit{tokyo}|\textit{7}}\\
\multicolumn{2}{c}{\textit{resto\_tokyo\_affordable\_vietnamese\_6stars}|\textit{vietnamese}|\textit{affordable}|\textit{tokyo}|\textit{6}}\\
\multicolumn{2}{c}{\textit{resto\_tokyo\_affordable\_vietnamese\_5stars}|\textit{vietnamese}|\textit{affordable}|\textit{tokyo}|\textit{5}}\\
\midrule
\midrule
\textbf{usr-1} & can you book a table in \textit{tokyo} with \textit{vietnamese} food in a \textit{affordable} price range ? \\
\textbf{sys-1} & what do you think of this option : \textit{resto\_tokyo\_affordable\_vietnamese\_8stars} ? \\
\textbf{usr-2} & no this does not work for me . \\
\textbf{sys-2} & what do you think of this option : \textit{resto\_tokyo\_affordable\_vietnamese\_7stars} ? \\
\textbf{usr-3} & do you have something else ? \\
\midrule
\textbf{Gold} & what do you think of this option : \textit{resto\_tokyo\_affordable\_vietnamese\_6stars}
 \\
\midrule
\midrule
\specialcell[t]{\textbf{Seq2Seq+Copy}} & what do you think of this option : what ?
 \\
\midrule
\textbf{Seq2Seq} & what do you think of this option : \textit{resto\_london\_moderate\_british\_2stars} ? \\

\midrule
\textbf{Mem2Seq} & what do you think of this option : \textit{resto\_tokyo\_affordable\_vietnamese\_5stars} ?
 \\
\midrule
\textbf{\sys\ } & what do you think of this option : \textit{resto\_tokyo\_affordable\_vietnamese\_6stars} ? \\
\bottomrule
\end{tabular}
\caption{An example of responses generated by \sys\ and baselines on bAbI dialog Task-5. This example is from the KA test set with 100\% unseen entities.}
\label{tab:task5-qualeval}
\end{table*}

\section{Related Work}
Compared to the traditional slot-filling based dialog  \cite{williams2007partially,wen2017network,williams2017hybrid}, 
end-to-end training methods (e.g., \cite{BordesW16}, this work) do not require handcrafted state representations and their corresponding annotations in each dialog. Thus, they can easily be adapted to a new domain.  We discuss end-to-end approaches along two verticals: 1) decoder: whether the response is retrieved or generated and 2) encoder: how the dialog history and KB tuples are encoded.

Most of the existing end-to-end approaches  {\em retrieve} a response from a pre-defined set \cite{BordesW16,liu2017gated,seo2016query}. These methods are generally successful when they have to provide boilerplate responses -- they cannot construct responses by using words in KB not seen during training. 
Alternatively, generative approaches are used where the response is {\em generated} one word at a time \cite{eric2017copy,mem2seq}. These approaches mitigate the unseen entity problem by incorporating the ability to copy words from the input \cite{vinyals2015pointer,gu2016incorporating}. The copy mechanism has also found success in summarization \cite{nallapati2016abstractive,see2017get} and machine translation \cite{ptr-unk}. \sys\ is also a copy incorporated generative approach.

For encoding, some approaches represent the dialog history as a sequence \cite{eric2017copy,ptr-unk}. Unfortunately, using a single long sequence for encoding also enforces an order over the set of KB tuples making it harder to perform inferencing over them. Other approaches represent the dialog context as a bag. Original Memory Networks \cite{BordesW16} and its extensions encode each memory element (utterance) as an average of all constituent words -- this cannot point to individual words, and hence cannot be used with a copy mechanism. Mem2Seq encodes each word individually in a flat memory. Unfortunately, this loses the contextual information around a word, which is needed to decipher an unseen word. In contrast, \sys\ uses a bag of sequences encoding, where KB tuples are a set for easier inference, and also each utterance is a sequence for effectively learning when to copy.

\section{Conclusions}
We propose \sys\ for training task-oriented dialog systems in an end-to-end fashion. \sys\ combines a novel bag of sequences memory for storing a dialog history and KB tuples, with a copy-augmented generative decoder to construct dialog responses. It augments standard cross entropy loss of a sequence decoder with an additional term to encourage the model to copy KB words. {\sc BoSs} memory and new loss term, in conjunction with a disentangle label dropout, enables the decoder to disentangle its language and knowledge models. 

\sys\ achieves the state of the art results on bAbI dialog dataset, outperforming existing models by 10 points or more in its OOV conditions. In the knowledge adaptability test, we find that \sys\ is highly robust to increasing the percentage of unseen entities at test time, suggesting a good language-knowledge disentanglement. Human evaluations show that \sys\ responses are highly informative and slightly more grammatical compared to baselines. We will release our code and all curated datasets for further research.



\section*{Acknowledgments}
We thank Danish Contractor, Gaurav Pandey and Sachindra Joshi for their comments on an earlier version of this work.  This work is supported by IBM AI Horizons Network grant, an IBM SUR award, grants by Google, Bloomberg and 1MG, and a Visvesvaraya faculty award by Govt. of India. We thank Microsoft Azure sponsorships, and the IIT Delhi HPC facility for computational resources.

\bibliography{naaclhlt2019}
\bibliographystyle{acl_natbib}

\appendix

\section{Two-Level attention on BoSs Memory}
\label{ssec:hierarattn}

\begin{figure*}
\centering
\includegraphics[width=0.8\textwidth]{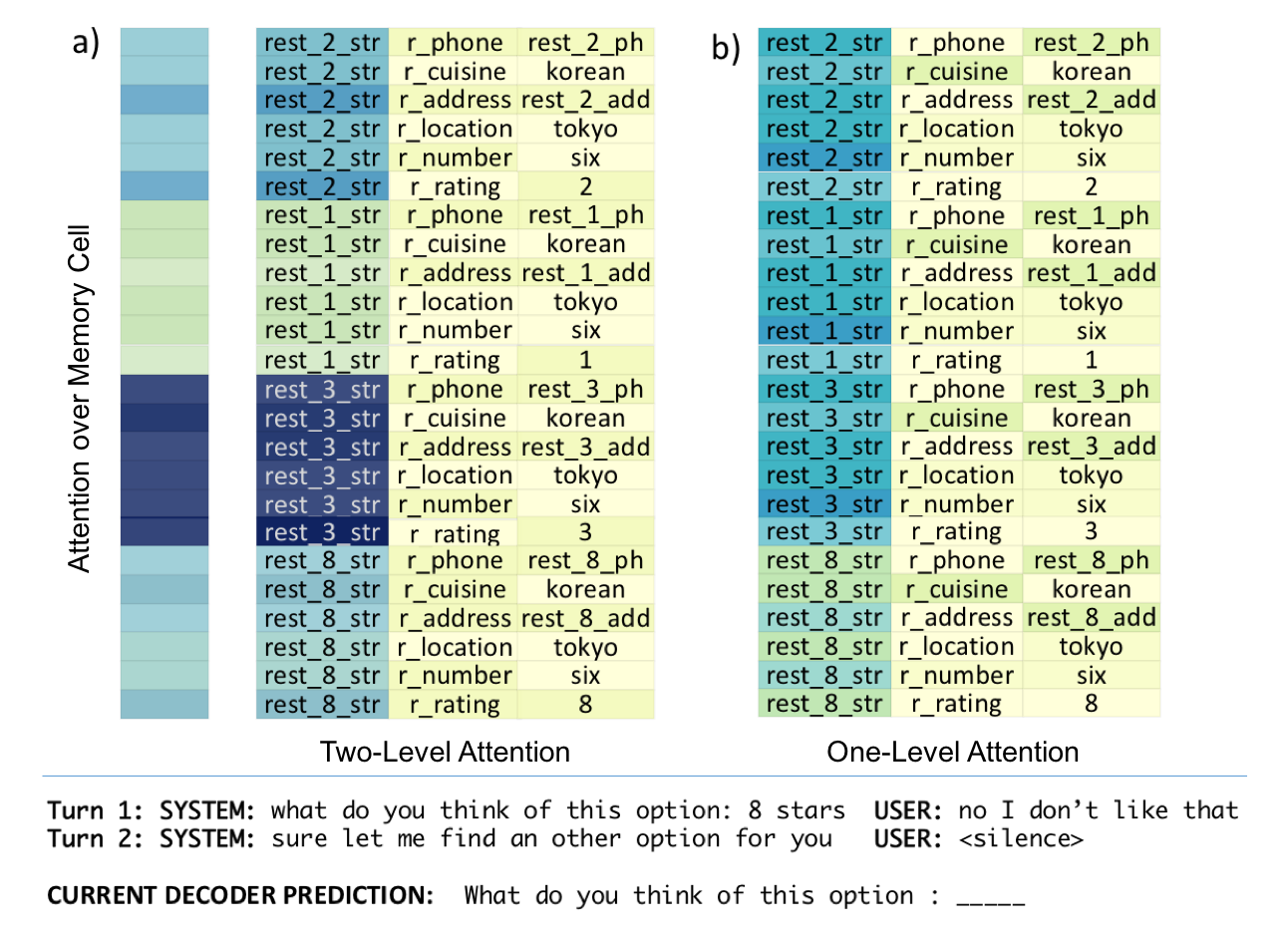}
\caption{Visualization of attention weights on selected portions of memory in (a) \sys\ with two-level attention vs (b) \sys\ with one-level attention}
\label{fig:attention}
\end{figure*}

To visualize the benefit of two-level attention used on {\sc BoSs} memory by the decoder, we compare attention weights for two models: our proposed \emph{two-level attention} and a variant with just \emph{one-level attention} (over all the words in the memory). In the example of a sample dialog from bAbI Task 3, shown in Figure \ref{fig:attention}, the decoder is aimed at predicting the second best restaurant \textit{3 stars}, given that the restaurant with rating \textit{8 stars} has already been suggested and rejected. We show attention only on the KB entries for brevity.

The models share some similarities in their distribution of attention. First, the attention weights are localized over the restaurant names, indicating the preference of the system to point to a specific restaurant. This is supported by the $g_s$ values, $3.14$ x $10^{-5}$ and $1.15$ x $10^{-4}$ for two-level attention and one-level attention respectively, i.e., both models prefer to copy rather than generate. Moreover, entries with the same restaurant name have similar attention weights, reflecting the robustness of the distribution.

We also observe that two-level attention is able to perform the difficult task of {\em sorting} the restaurant entries based on decreasing order of rating (number of stars). It gives more weight to entries with a high rating 
(\textit{3 stars} $>$ \textit{2 stars} $>$ \textit{1 star})
and suppresses the weights of any previously suggested restaurant.

The attention over memory cells provides \sys\ with the ability to infer over multiple sets of tuples. The ability to sort the restaurants and reject a previously seen restaurant can be observed by the attention heat map of Memory cells. Attention over tokens on the other hand can push the attention weights towards either the subject or object in the KB tuple, based on the query's request. Thus using both in conjunction helps \sys\ perform significantly better than the baselines and illustrates the importance of the {\sc BoSs} memory in comparison to a flat memory layout.

\section{Reproducibility}
\label{sec:reproduce}
We list out the complete set of hyperparameters used to train \sys\ for the various datasets in Table \ref{tab:params}. Our code will be made publicably accessible for future research purposes. Our trained models and evaluation scripts will also be provided. We will also make our end-to-end reconstruced Camrest dataset along with our whole batch of knowledge adaptability test sets available.

\begin{table*}[ht]
\centering
\footnotesize
\begin{tabular}{c|ccccc}
\toprule
\textbf{Task} & \textbf{Learning Rate} & \textbf{Hops} &  \textbf{Embedding Size} & \textbf{Disentangle Loss Weight} & \textbf{DLD}\\
\midrule
T1 & 0.001 & 1 & 128 & 1.0 & 0.2 \\
T2 & 0.001 & 1 & 128 & 1.0 & 0.2 \\
T3 & 0.0005 & 3 & 128 & 1.5 & 0.2 \\
T4 & 0.001 & 1 & 128 & 1.0 & 0.2 \\
T5 & 0.0005 & 3 & 256 & 1.0 & 0.2 \\
CamRest & 0.0005 & 6 & 256 & 1.0 & 0.2 \\
SMD & 0.0005 & 3 & 256 & 1.0 & 0.1 \\
\bottomrule 
\end{tabular}
\caption{The hyperparameters used to train \sys\ on the different datasets}. 
\label{tab:params}
\end{table*}

\section{Example Predictions of \sys\ and Baselines}
\label{sec:examples}
Examples from SMD is shown in Table \ref{tab:smd0} respectively. Examples from KA test set with percentage of unseen entites set to 50 from CamRest and SMD are shown in Table \ref{tab:cam50} and Table \ref{tab:smd50} respectively. Examples from KA test set with percentage of unseen entites set to 100 from bAbI dialog Task 1 is shown in Table \ref{tab:t1_100}.

\input{task1_examples.tex}
\input{camrest_examples.tex}
\input{smd_examples.tex}

\section{Dataset Preprocessing and Faults}
\label{sec:preprocess}
\subsection{Mem2Seq Preprocessing}
\label{sec:prep_mem}

\begin{figure*}[ht]
\centering
\includegraphics[width=0.8\textwidth]{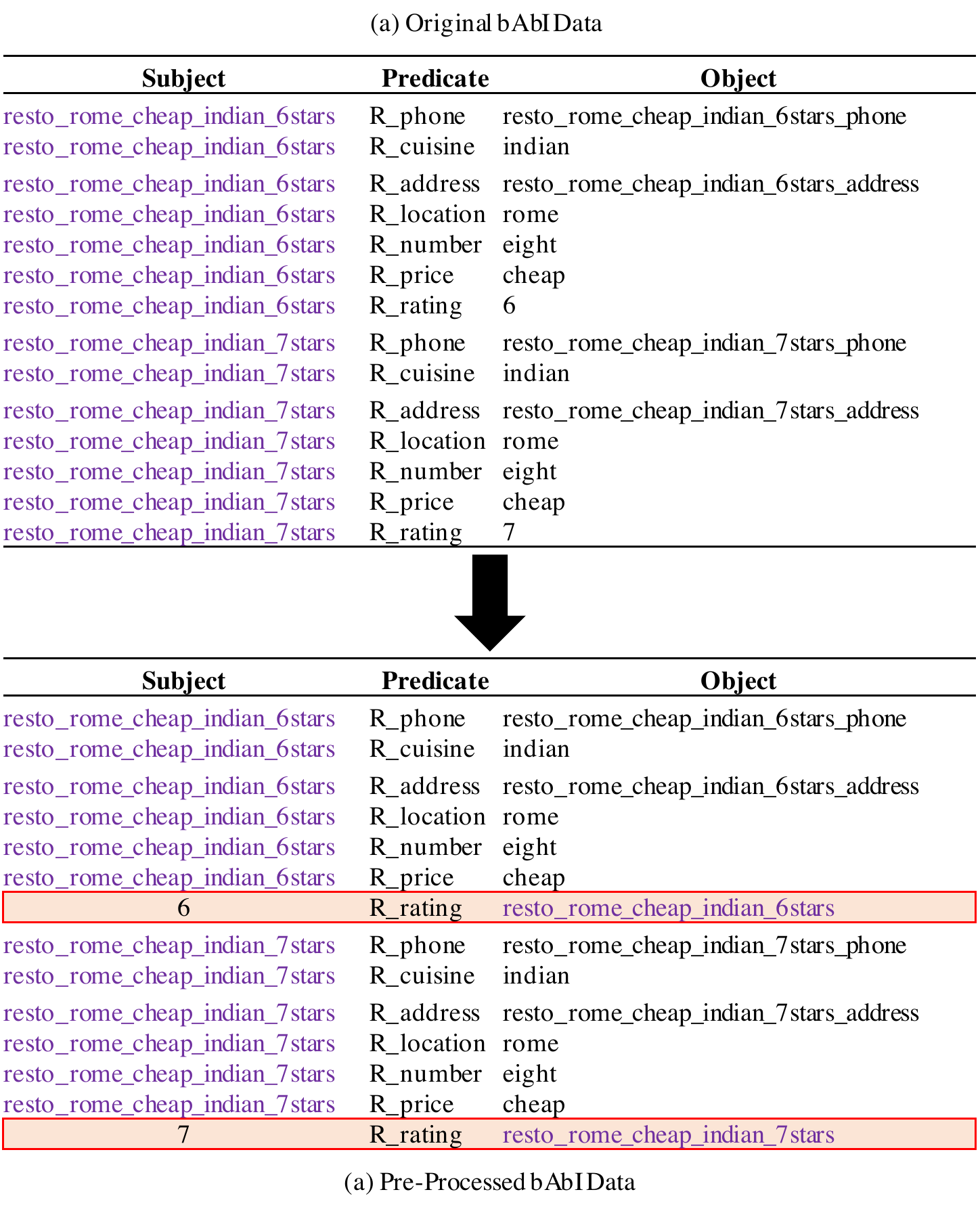}
\caption{Pre-processing of bAbI dialog data used in Mem2Seq paper}
\label{fig:prebabi}
\end{figure*}

\begin{figure*}[ht]
\centering
\includegraphics[width=0.8\textwidth]{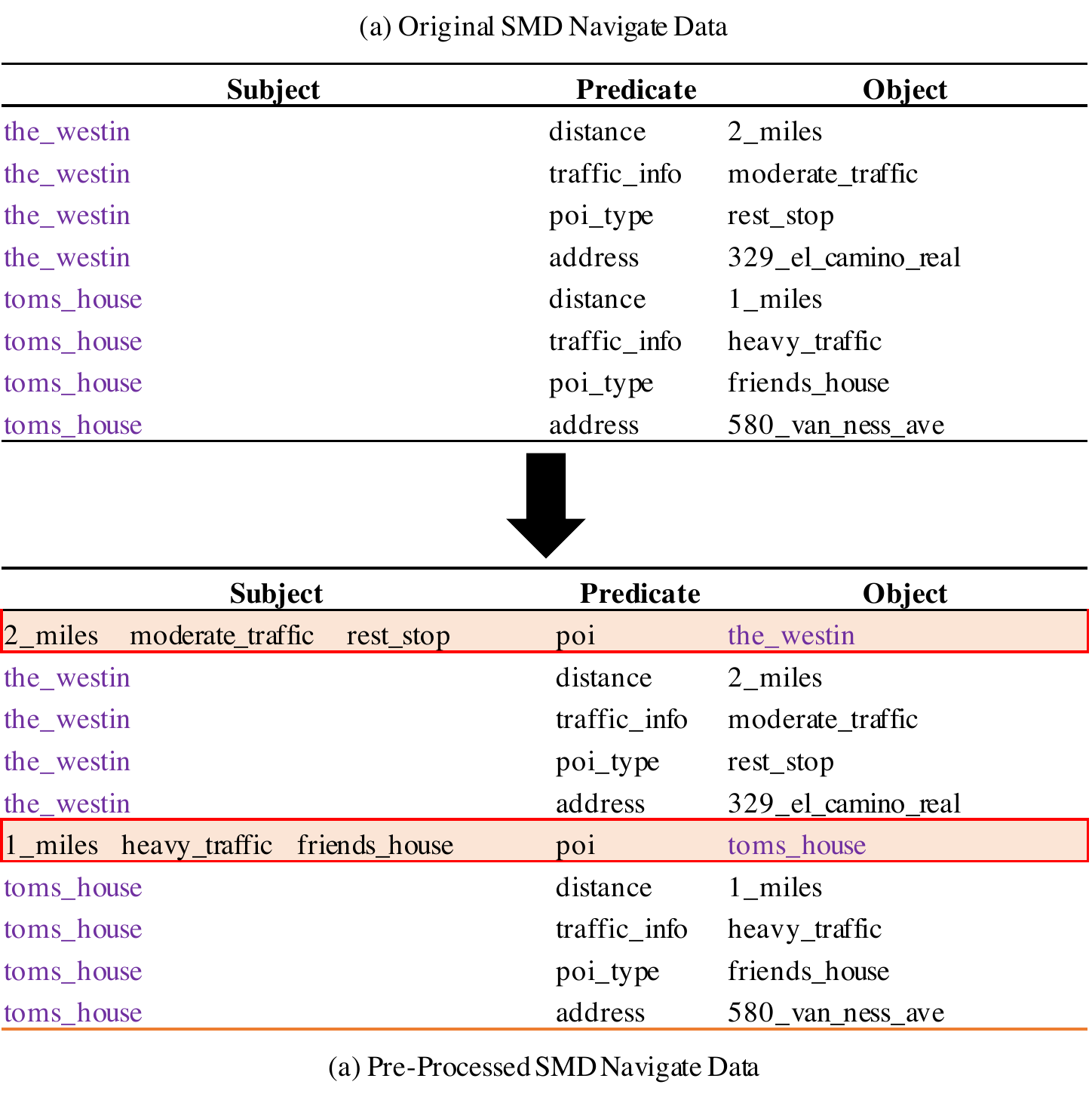}
\caption{Pre-processing of SMD Navigate data used in Mem2Seq paper}
\label{fig:presmd}
\end{figure*}

Mem2Seq paper used the following pre-processing on the data:
\begin{enumerate}
    \item The subject (restaurant name) and object (rating) positions of the rating KB tuples in bAbI dialogs are flipped, while the order remains the same for other tuples remains the same. This pre-processing is illustrated in Figure \ref{fig:prebabi}
    \item an extra fact was added to the navigation tasks in In-Car Assistant with all the properties (such as distance, address) combined together  as the subject and \textit{poi} as the object. This pre-processing is illustrated in Figure \ref{fig:presmd}
\end{enumerate}
The pre-processing has major impact on the performance of  Mem2Seq, as it can only copy objects of a KB tuple, while the subject and relation can never be copied.

\subsection{bAbI Dataset Faults}
\label{sec:fault}
The KB entities present in validation and non-OOV test sets for task 3 and 4 do not overlap with those in the train set. This effectively means that non-OOV and OOV test conditions are the same for tasks 3 and 4. This explains the low performance of baseline models on task 3 and 4 non-OOV test sets.

\section{AMT Setup}
\label{sec:amt_setup}
\begin{figure*}[h]
    \centering
    \subcaptionbox{\label{sfig:testa}}{
    \includegraphics[width=0.9\textwidth]{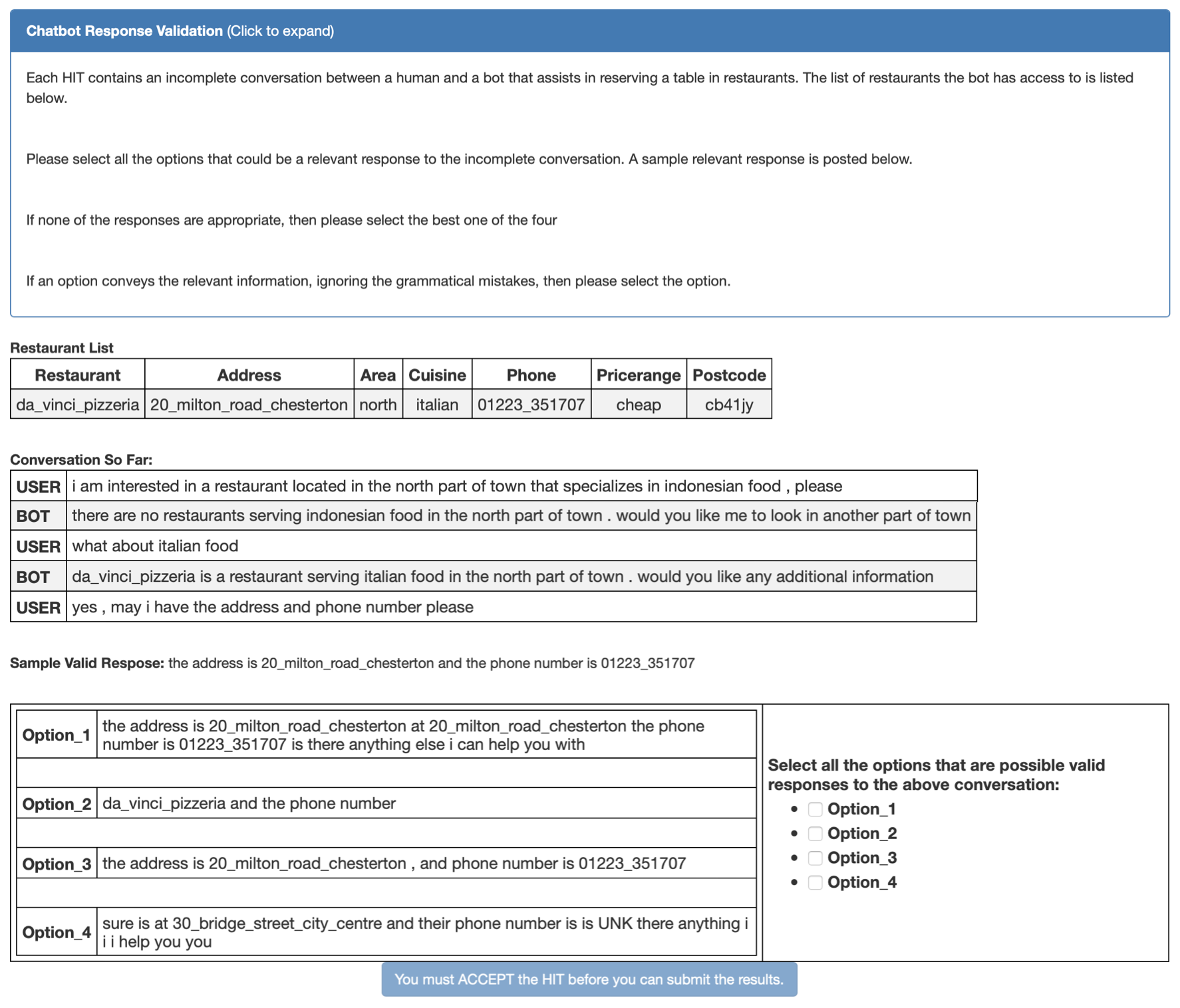}}
    \subcaptionbox{\label{sfig:testb}}{\includegraphics[width=0.9\textwidth]{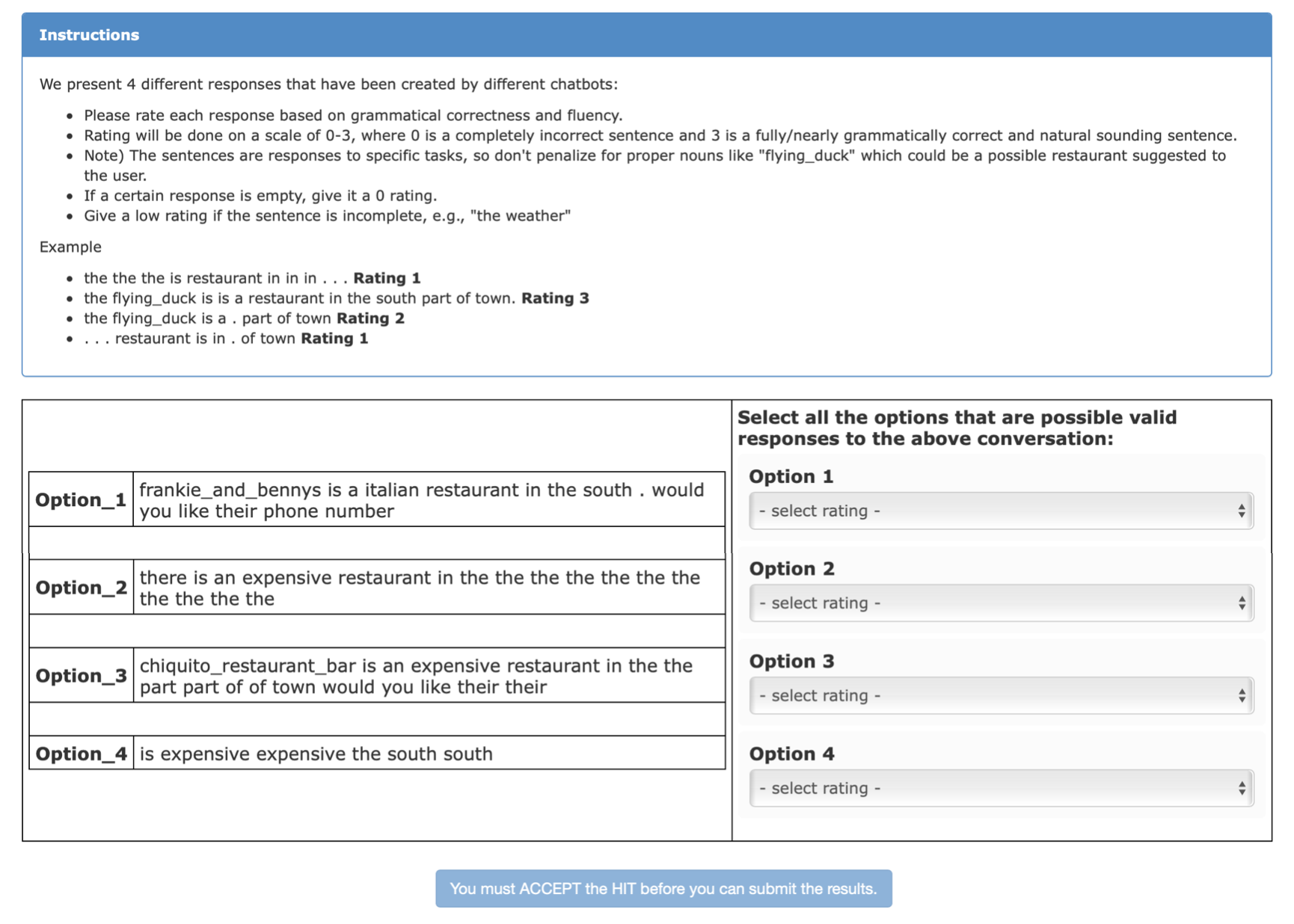}}
    \caption{A sample HIT on Amazon Mechanical Turk to (a) validate useful responses based on the given dialog context, and (b) validate grammatical correctness of different responses on a scale of 0-3}
    \label{fig:amt_rel}
\end{figure*}

\noindent \textbf{Response Relevance Test} 
We show a sample of an Human Intelligence Task (HIT) on Amazon Mechanical Turk in Figure \ref{sfig:testa}. We randomize the responses generated by the three baseline models and \sys\ on the same dialog and ask the user to tick all those response options that seem to capture the relevant information of the given sample response. A total of 200 such annotations were collected for Camrest and SMD each.

\noindent \textbf{Response Grammar Test}
We show a sample of an Human Intelligence Task (HIT) on Amazon Mechanical Turk in Figure \ref{sfig:testb}. We randomize the responses generated by the three baseline models and \sys\ on the same dialog and ask the user to rate each response based on the grammatical correctness and natural flow of the sentence. The rating ranges from 0-3 where 0 being the worst and 3 being the best. Note) the sentences were not asked to be rated with respect to each other, but instead as individual occurrences. A total of 200 such annotations were collected for Camrest and SMD each.

\section{Multi-Hop vs 1-Hop Encoders}
Table \ref{tab:ablationhop} shows the performance of bAbI tasks and CamRest on two \sys\ encoder settings. Multi-hops in encoder helps in bAbI task 3 and 5, as they require inferencing over the KB tuples (sorting restaurants by rating) to recommend a restaurant. We also see substantial improvements on CamRest in both BLEU and entity F1 metric.

\begin{table*}
\centering
\footnotesize
\begin{tabular}{l|ccccc|ccccc|cc}
\toprule
   & \multicolumn{5}{c|}{\textbf{bAbI Dialog Tasks}} & \multicolumn{5}{c|}{\textbf{bAbI Dialog Tasks (OOV)}}  & \multicolumn{2}{c}{\textbf{CamRest}} \\ \cmidrule{2-6} \cmidrule{7-11} \cmidrule{12-13}
    & T1  & T2  & T3   & T4   & T5   & T1 & T2 & T3 & T4 & T5 & BLEU        & Ent. F1       \\ \midrule
\sys\ with 1-Hop Encoder & 100 & 100 & 92.3 & 100 & 90.5 & 100 & 100 & 91.4 & 100 & 89 & 10.5 & 36.9 \\
\sys\ with Multi-Hop Encoder & 100 & 100 & 95.2 & 100  & 97.3 & 100    & 100    & 95.7   & 100    & 91.7   & 15.2        & 43.1    
\\ \bottomrule
\end{tabular}
\caption{Ablation study: impact of hops in \sys\ encoder }
\label{tab:ablationhop}
\end{table*}

\end{document}

%% file: weld-defns.tex
\newcommand{\?}{\mbox{?}}

\newcommand{\bi}{\begin{list}{$\bullet$}{
    \setlength{\leftmargin}{1.5 em}
    \setlength{\itemsep}{0 pt}
    \setlength{\topsep}{3 pt}
    \setlength{\parsep}{3 pt}
    \setlength{\partopsep}{0 pt}
    \setlength{\labelwidth}{1 em}
    \setlength{\labelsep}{0.5 em}
    \setlength{\parskip}{0cm}  }}
\newcommand{\ei}{\end{list}}

\newcommand{\BE}{\begin{enumerate}}
\newcommand{\EE}{\end{enumerate}}

\newcommand{\initab}{                           
\begin{tabbing}
XXX \= XXXX \= \kill
}
\newcommand{\begpub}{
\begin{quotation}
\noindent
}

\newcommand{\finpub}{
\end{quotation}
}

%% file: task1_examples.tex
\begin{table*}
\centering
\small
\begin{tabular}{c|l}

\toprule
\midrule
\textbf{usr-1} &good morning \\
\textbf{sys-1} & hello what can i help you with today \\
\textbf{usr-2} & can you book a table with \textit{korean} cuisine for \textit{five} in \textit{beijing} \\
\textbf{sys-2} & i ' m on it . which price range are looking for ? \\
\textbf{usr-3} & i am looking for a \textit{overpriced} restaurant \\
\midrule
\textbf{Gold} & api\_call \textit{korean beijing five overpriced}
 \\
\midrule
\midrule
\specialcell[t]{\textbf{Seq2Seq+Copy}} & api\_call \textit{korean beijing five overpriced}
 \\
\midrule
\textbf{Seq2Seq} & api\_call \textit{french london four moderate} \\

\midrule
\textbf{Mem2Seq} & api\_call \textit{overpriced korean korean overpriced}
 \\
\midrule
\textbf{\sys\ } & api\_call \textit{korean beijing five overpriced} \\
\bottomrule
\end{tabular}
\caption{Example from bAbI dialog Task 1 with 100\% OOV.}
\label{tab:t1_100}
\end{table*}

%% file: camrest_examples.tex
\begin{table*}
\centering
\small
\begin{tabular}{c||p{13cm}}
\toprule
\multicolumn{2}{c}{\textbf{KB (restaurant|cuisine|address|phone|postcode)}} \\
\multicolumn{2}{c}{\textit{prezzo}|\textit{ethiopian}|\textit{22\_atlantis\_road}|\textit{01799\_521260}|\textit{cb30ad}}\\
\midrule
\midrule
\textbf{usr-1} &i want a \textit{moderately\_expensive ethiopian} restaurant in the \textit{south} side of town \\
\textbf{sys-1} & i ' m sorry . i can not find a \textit{moderately\_expensive ethiopian} restaurant in the \textit{south} side of town \\
\textbf{usr-2} & i would like to find a \textit{moderately} priced restaurant in the \textit{west} part of town \\
\textbf{sys-2} & \textit{prezzo} is a moderately priced \textit{ethiopian} restaurant in the \textit{west} part of town . there is also \textit{indian} and \textit{british} food available in that area and price range . would you like more info ? \\
\textbf{usr-3} & yes can i get thier address and postcode ? \\
\midrule
\textbf{Gold} & of course , \textit{prezzo} ' s address is \textit{22\_atlantis\_road} and postcode is \textit{cb30ad}
 \\
\midrule
\midrule
\specialcell[t]{\textbf{Seq2Seq+Copy}} & postcode a is is \textit{01223350688} UNK there get i for for today
 \\
\midrule
\textbf{Seq2Seq} & the address is \textit{83\_regent\_street} and their postcode is \textit{cb58aq} \\

\midrule
\textbf{Mem2Seq} & the phone address is \textit{cb21uj}
 \\
\midrule
\textbf{\sys\ } & the address is \textit{22\_atlantis\_road at 22\_atlantis\_road} the postcode is \textit{cb30ad} is there anything else i can help you with \\
\bottomrule
\end{tabular}
\caption{Example from Camrest with 50\% OOV. The OOV entities present in the dialog are \{\textit{ethiopian}, \textit{22\_atlantis\_road}\}}
\label{tab:cam50}
\end{table*}

%% file: smd_examples.tex
\begin{table*}
\centering
\small
\begin{tabular}{c|p{13cm}}

\toprule
\multicolumn{2}{c}{\textbf{KB (location|distance|traffic info|address)}} \\
\multicolumn{2}{c}{\textit{teavana}|\textit{4\_miles}|\textit{no\_traffic}|\textit{145\_amherst\_st}}\\
\multicolumn{2}{c}{\textit{cafe\_venetia}|\textit{5\_miles}|\textit{moderate\_traffic}|\textit{269\_alger\_dr}}\\
\multicolumn{2}{c}{\textit{town\_and\_country}|\textit{3\_miles}|\textit{moderate\_traffic}|\textit{383\_university\_ave}}\\
\midrule
\midrule
\textbf{usr-1} & where can i get tea ? \\
\textbf{sys-1} & there is 2 tea shops near you \textit{teavana} is \textit{4\_miles} away and \textit{cafe\_venetia} is \textit{5\_miles} away which one do you want ? \\
\textbf{usr-2} & which one is the quickest is there any traffic ? \\
\midrule
\textbf{Gold} & \textit{teavana} is the fastest journey because there is \textit{no\_traffic} on the way there . \textit{cafe\_venetia} has \textit{moderate\_traffic} should we go to \textit{teavana}
 \\
\midrule
\midrule
\textbf{Seq2Seq+Copy} & there is
 \\
\midrule
\textbf{Seq2Seq} & there is \textit{moderate\_traffic} on the way to \textit{palo\_alto\_medical\_foundation} \\

\midrule
\textbf{Mem2Seq} & there is \textit{moderate\_traffic} on our way
 \\
\midrule
\textbf{\sys\ } & \textit{teavana} there is a \textit{moderate\_traffic} nearby but i might find another route to t\textit{own\_and\_country} \textit{town\_and\_country} address is \textit{383\_university\_ave} i sent it on your screen \\
\bottomrule
\end{tabular}
\caption{Example from SMD}
\label{tab:smd0}
\end{table*}

\begin{table*}
\centering
\small
\begin{tabular}{c|p{13cm}}

\toprule
\multicolumn{2}{c}{\textbf{KB (location|day|weather)}} \\
\multicolumn{2}{c}{\textit{mountain\_view}|\textit{sunday}|\textit{foggy}}\\
\multicolumn{2}{c}{\textit{mountain\_view}|\textit{monday}|\textit{overcast}}\\
\multicolumn{2}{c}{\textit{pittsburgh}|\textit{sunday}|\textit{overcast}}\\
\multicolumn{2}{c}{\textit{pittsburgh}|\textit{monday}|\textit{warm}}\\
\multicolumn{2}{c}{\textit{downtown\_chicago}|\textit{sunday}|\textit{foggy}}\\
\multicolumn{2}{c}{\textit{downtown\_chicago}|\textit{monday}|\textit{warm}}\\
\midrule
\multicolumn{2}{c}{\textit{today}|\textit{monday}}\\
\midrule
\midrule
\textbf{usr-1} &is the weather \textit{cloudy} right now ? \\
\textbf{sys-1} & what city do you want the weather for ? \\
\textbf{usr-2} & \textit{pittsburgh} \\
\midrule
\textbf{Gold} & there are no clouds in \textit{pittsburgh} right now
 \\
\midrule
\midrule
\textbf{Seq2Seq+Copy} & which is no
 \\
\midrule
\textbf{Seq2Seq} & it is currently raining in \textit{san\_francisco} right now \\

\midrule
\textbf{Mem2Seq} & it will not be cloudy in in
 \\
\midrule
\textbf{\sys\ } & it will be \textit{cloudy} on \textit{sunday} in \textit{pittsburgh} \\
\bottomrule 
\end{tabular}
\caption{Example from SMD with 50\% OOV. The OOV entity present in the dialog is \{\textit{pittsburgh}\}}
\label{tab:smd50}
\end{table*}